\documentclass[sigconf,screen]{acmart}
\renewcommand\footnotetextcopyrightpermission[1]{} 
\makeatletter
\renewcommand\@formatdoi[1]{\ignorespaces}
\makeatother
\usepackage{svg}
\usepackage{amsmath}
\usepackage{footmisc}
\usepackage{mathrsfs}
\usepackage{adjustbox}
\usepackage{tabularx}
\usepackage{booktabs}
\usepackage{multirow}
\usepackage{flushend}
\usepackage{breqn} 
\usepackage{mathtools}

\newcolumntype{m}[1]{>{\centering\arraybackslash}p{#1}}
\usepackage{subcaption}
\usepackage{balance}
\usepackage{pifont}
\newcommand{\cmark}{\ding{51}}%
\newcommand{\xmark}{\ding{55}}
\definecolor{OliveGreen}{rgb}{0,0.6,0}
\definecolor{SoftRed}{rgb}{1,0.2,0.2}
\usepackage{xcolor}
\newcommand{\capitalhyphen}{\raisebox{0.24ex}{\resizebox{0.4em}{\height}{-}}\kern-0.07em}
\usepackage{color}
\usepackage{caption}
\usepackage{epsfig}
\usepackage{enumitem}
\usepackage{mathtools}
\usepackage{epsfig}
\usepackage{graphicx}
\usepackage{placeins} 
\usepackage{footnote} 

\begin{document}
\fancyhead{}

\title[MUGL]{MUGL: Large Scale Multi Person Conditional Action Generation with Locomotion}

\author{Shubh Maheshwari}
\email{maheshwarishubh98@gmail.com}
\affiliation{%
  \institution{CVIT, IIIT Hyderabad}
  \city{Hyderabad 500032}
  \country{INDIA}
}
\author{Debtanu Gupta}
\authornote{Equal contribution}
\email{debtanu.gupta@research.iiit.ac.in}
\affiliation{%
  \institution{CVIT, IIIT Hyderabad}
  \city{Hyderabad 500032}
  \country{INDIA}
}
\author{Ravi Kiran Sarvadevabhatla}
\email{ravi.kiran@iiit.ac.in}
\affiliation{%
 \institution{CVIT, IIIT Hyderabad}
 \city{Hyderabad 500032}
 \country{INDIA}
 }
 
\renewcommand{\shortauthors}{Maheshwari et al.}

\begin{abstract}
   We introduce MUGL, a novel deep neural model for large-scale, diverse generation of single and multi-person pose-based action sequences with locomotion. Our controllable approach enables variable-length generations customizable by action category, across more than $100$ categories. To enable intra/inter-category diversity, we model the latent generative space using a Conditional Gaussian Mixture Variational Autoencoder. To enable realistic generation of actions involving locomotion, we decouple local pose and global trajectory components of the action sequence. We incorporate duration-aware feature representations to enable variable-length sequence generation. We use a hybrid pose sequence representation with 3D pose sequences sourced from videos and 3D Kinect-based sequences of NTU-RGBD-120. To enable principled comparison of generation quality, we employ suitably modified strong baselines during evaluation. Although smaller and simpler compared to baselines, MUGL provides better quality generations, paving the way for practical and controllable large-scale human action generation. 
\end{abstract}

\begin{teaserfigure}
   \centering
     \includegraphics[width=\textwidth]{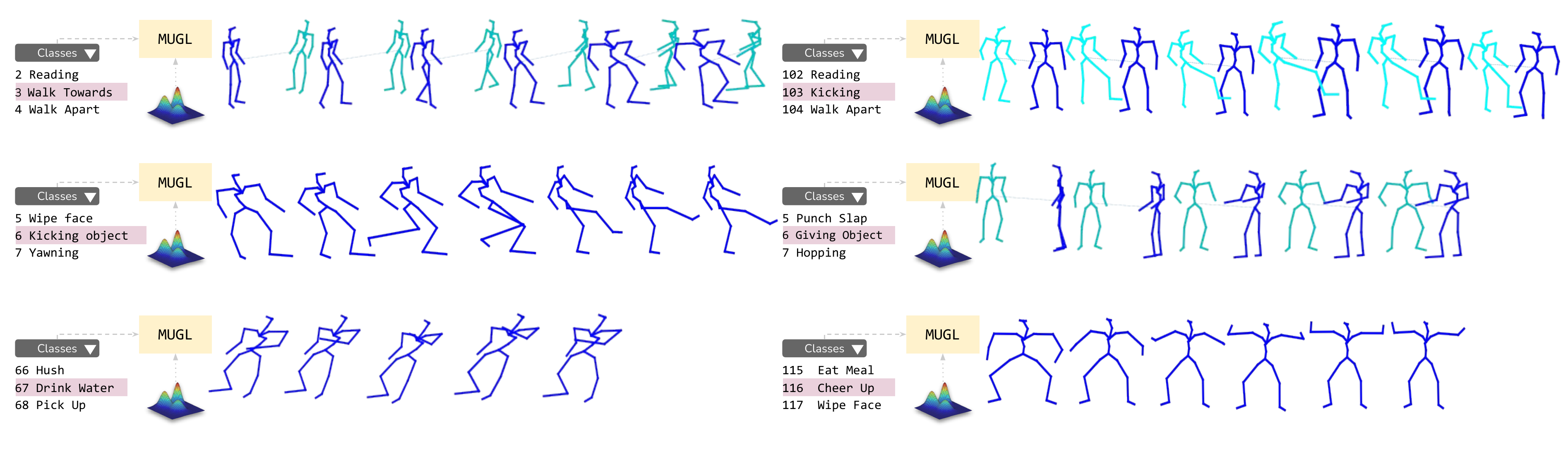}
     \captionof{figure}{Examples of single and multi-person human pose action sequences generated by our model, MUGL. The user can select from among a large number of action classes ($\mathbf{120}$). The changing length of dotted line connecting kinematic tree root joints of individuals in two-person actions indicates change in their relative global positions. Note the distinction from single person actions where individual frames correspond to timesteps, i.e. global displacements are absent. Also note the variable length of sequences generated. Additional results in Fig.~\ref{fig:generations}.}
     \label{fig:introfig}
\end{teaserfigure}


\maketitle


\section{Introduction}
\label{sec:intro}

The ability to synthesize novel and diverse human motion at scale while satisfying structural and kinematic body constraints has significant applications in animation, robotics and human-object interaction. Over the years, various approaches have been proposed including physics based simulation~\cite{deepMimic}, key-framing~\cite{motion_graph} and data-driven methods~\cite{temploral_rbms,recurent_rbms}. Apart from pixel-based video data, the availability of reliable motion capture systems~\cite{zhang2012microsoft,h36m,cmu_mocap} has enabled pose-based human action synthesis~\cite{holden_phase,aclstm,quaternet}. Unlike videos, pose-based action representations and data have the significant advantage of decoupling the action-centric aspects from distracting or privacy-violating aspects (e.g. entity appearance, background).

However, pose-based human action synthesis is very challenging because of spatiotemporal correlation arising from body joint movement and the inter/intra category variety among actions. The challenge is further amplified by the presence of multiple actors. Due to the resulting complexity and lack of large-scale datasets~\cite{h36m,cmu_mocap,kit_motion}, many approaches  focus on synthesis of small action sets of fixed duration and simple, single-person actions~\cite{holden_phase,aclstm,quaternet}. 

In this backdrop, the availability of the large-scale NTU-RGBD dataset~\cite{ntu-120} is a promising development. The dataset contains $120$ diverse single, multi-person activities performed indoors by a large subject pool and captured from multiple viewpoints using Kinect RGBD cameras. However, a fundamental bottleneck exists at the raw data stage itself. For a number of instances and action classes (e.g. `handshake', `pushing'), the Kinect-based 3D joint data is temporally incoherent and noisy. Na\"ively using this data for training would cause generative models to learn sub-optimal representations and result in unrealistic, poor-quality generations. Given recent advances, utilizing 3D human pose estimated directly from RGB videos is a promising alternative which addresses some issues mentioned above. Therefore, we source 3D human poses from RGB videos across all the $120$ action classes of the NTU-RGBD dataset. In doing so, we obtain a large-scale, good quality hybrid pose sequence representation which inherits advantages of the original viz. access to global trajectory (from Kinect-based sequences), large number of action classes, diversity in viewpoints and variety in action dynamics. In addition, the controlled indoor setting enables a principled comparison of generative model quality. 

Despite the availability of large-scale datasets such as NTU-RGBD-120, existing approaches have not shown results beyond a small number of categories and single-person actions~\cite{yu2020structureaware,action2motion}. The absence of locomotion with respect to global frame makes these methods unproductive for actions such as `walking' or `running'. Also, actions involving locomotion and multi-person interaction (e.g. `kicking') cannot be modelled properly. In addition, existing approaches generate fixed-length sequences causing certain generated actions (e.g. `throw', `salute', `take off headphone') to appear unnatural. 

To address these shortcomings, we propose MUGL, a novel generative model which enables large-scale diverse generation of single \textit{and multi-person} human actions of \textit{variable duration}, \textit{with locomotion}. Notably, we accomplish controllable generation with a single, unified model. Our contributions are summarized below:
\begin{itemize}
    \item We propose MUGL, a novel efficient deep network for large-scale controllable generation of  multi-person human action sequences with variability in action duration and with locomotion (Sec.~\ref{sec:architecture}). 
    \item We introduce a hybrid pose sequence representation with 3D pose sequences sourced from videos and 3D Kinect-based sequences of NTU-RGBD-120 (Sec.~\ref{sec:ntuvibe120}).
    \item MUGL outperforms strong baselines and generates visibly more realistic sequences for \textbf{all} $\mathbf{120}$ \textbf{categories} of NTU-120 (Sec.~\ref{sec:results}).
\end{itemize}

Additional details can be found in our project page \url{skeleton.iiit.ac.in/mugl}.

\section{Related Work}

\noindent \textbf{Action prediction:} A small initial action sequence is used to condition the generation of the full version in action prediction task. One set of approaches employ a sequence-to-sequence paradigm to predict joints~\cite{Fragkiadaki2015RecurrentNM,Battan_2021_WACV}  or joint velocities~\cite{martinez2017human}. Other approaches use adversarial generative models~\cite{hpgan,bihmpgan} and conditioned autoregressive models~\cite{holden_phase} to synthesize human leg motion. Gao et al.~\cite{skelnet} also use an autoregressive setup involving disjoint part grouping of skeleton joints. However, they train separate models for each action category. To produce sequences, Battan et al.~\cite{Battan_2021_WACV} use a two stage approach involving sparse and dense motion  prediction. In general, however, end-to-end action synthesis is more challenging due to the absence of input priming used for action prediction.

\noindent \textbf{Traditional Action Generation:} Some early works model motion sequences as a directed graph of pose subsequences sampled from motion capture corpus~\cite{motion_graph,motion_examples,motion_annotations}. These methods are storage-intensive and generalization is limited to producing memorized sequences. Another set of approaches use Restricted Boltzmann Machines (RBM)~\cite{rbm} for probabilistic human motion generation~\cite{temploral_rbms,recurent_rbms}. Though possessing the advantage of learning a probabilistic distribution, these methods involve conditioning on proper choice of initial state which inhibits scaling to complex, large number of activities.   

\noindent \textbf{Modern Action Generation:} Peng et al.~\cite{deepMimic} use reinforcement learning to create realistic motion clips imitating a broad range including locomotion acrobatics. Peng et al.~\cite{10.1145/3272127.3275014} further extend the earlier work using deep pose estimation to incorporate publicly available RGB videos. However, these works mainly focus on using a single motion clip for action sequence generation. Yan et al.~\cite{CSGCN} propose to generate the action sequence using a graph convolution based GAN model with generation conditioned on a latent vector sampled from a Gaussian process. Yu et al.~\cite{yu2020structureaware} propose a novel self-attention based GCN method with category conditioning of a GAN network. However, the focus is on generation of 2-D skeletons and only $10$ action categories are considered. Guo et al.~\cite{action2motion} propose a conditional VAE and also use RGB-based skeletons. However, their generations are confined to a small number of single-person action categories ($13$). 
 
In a slightly different paradigm, some approaches use natural language action descriptions to condition the generation process. Chaitanya et al.~\cite{language2pose}'s approach involves learning a joint embedding space for both action and language representation. However, the model generates a single sample for a given action description, i.e. stochasticity is absent. Xiao et al.~\cite{DVGAN} adopt a similar approach but within a GAN-based setting which enables stochastic sequence generation. Hyemin et al.~\cite{text2action} propose an attention-based sequence to sequence generator confined to upper body parts.

\noindent \textbf{Action sequence representation:} Many methods use joint 3-D positions to represent human skeleton pose~\cite{skelnet,CSGCN,language2pose,text2action}. But this representation has a disadvantage of non-constrained bone length and motion beyond normal articulation range. To utilize the advantage of constrained bone length, some methods represent pose using joint rotations expressed as quaternions~\cite{quaternet}. The discontinuous nature of these representations, however, results in sub-optimal pose embeddings. To overcome this, we employ a continuous 6-D representation for joint rotation (Sec.~\ref{sec:skelerotationrep}).  

\noindent \textbf{Action sequence embedding:} Since human actions can be viewed as a sequence of human poses, many approaches~\cite{language2pose,skelnet} use Recurrent Neural Networks(RNN) to obtain action sequence embeddings. However, RNN-based methods often fail to model temporal structures at multiple scales and in generating long motion sequences. A recent group of approaches view the action as a spatiotemporal graph and apply graph convolutions to represent the same~\cite{yu2020structureaware,CSGCN}. However, these methods are computationally expensive, even for a small number of action classes. We use Convolutional Neural Networks(CNN) in our approach which are computationally efficient for representing the spatiotemporal structure of the action sequence. 

\noindent \textbf{Locomotion:} Estimation of locomotion is crucial for tasks such as multi-person interaction, object interaction, path planning. Cao et al.~\cite{caoHMP2020} use 2D pose histories and a scene image to plan a path towards reaching each goal. Habibie et al. ~\cite{vaelstm} propose a VAE conditioned on a user-supplied control signal (desired trajectory). Similarly Holden et. al. ~\cite{holden_2016} use a CNN autoencoder to create a low dimension representation and user-provided instructions to edit and synthesize long range motion. These approaches do not allow generation controllable by user at action category level. Phase functional neural networks~\cite{holden_phase} use cyclic functions to compute weights of the network during inference for high quality, real time character control. These methods are generally confined to only locomotion classes such as walking, running and terrain traversal and are unable to scale to large number of categories, including non-locomotion action classes. 

Overall, unlike our work, existing methods do not cover a large number of action classes, do not produce variable-length sequences and do not generate multi-person activity sequences with relative global displacements.

\section{Preliminaries}

\noindent \textbf{Problem formulation:} An action sequence $\mathcal{X}$ is associated with a class label $c \in \mathcal{C}$ and can involve up to $P$ individuals for up to $T$ timesteps. Formally, we denote the sequence as $\mathcal{X} = \{[X^{(1)}, X^{(2)}, \ldots \\ X^{(p)}]_t\}, 1 \leqslant p \leqslant P$ and $1 \leqslant t \leqslant T$.  $[X^{(i)}]_t$ represents the $J$-joint pose configuration of $i$ -th person at time step $t$ within the global coordinate frame. Note that the number of people involved in an action ($P$) and number of time steps can vary across action classes. Our goal is a model which stochastically generates a variable length action sequence $\mathcal{X}$ conditioned on class label $c$. 

In our setting, the number of action classes $|\mathcal{C}|$ in NTU dataset is $120$, maximum number of people in actions $P$ is $2$, number of joints $J$ is $24$ and timesteps $T$ is $64$.

\subsection{Action Sequence Representation}
\label{sec:skelerotationrep}

\begin{figure*}[!t]
    \includegraphics[width=\textwidth]{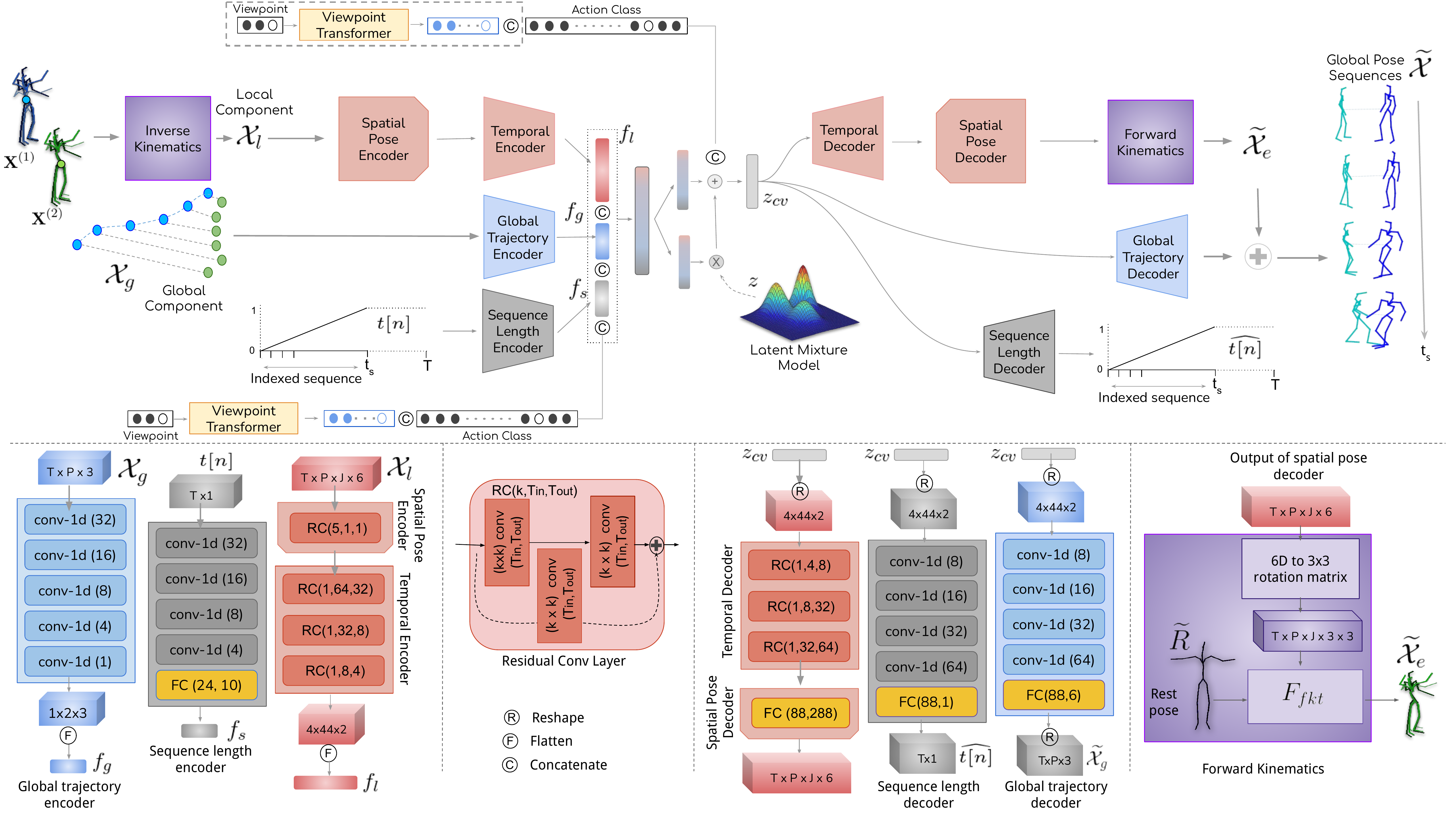}
    \caption{Architecture of MUGL. The green and blue circular markers at torso joint on $\mathbf{X}^{(1)},\mathbf{X}^{(2)}$ indicate the shared origin of action's local component. The dotted boundary around transformed viewpoint indicates that a fixed viewpoint is used during inference. During training, the input sequence's viewpoint is utilized. Refer to Sec.~\ref{sec:architecture} for details.}
    \label{fig:architecture}
\end{figure*}

Apart from actions performed in-place (without locomotion), our dataset includes actions involving locomotion and relative global motion involving multiple people. To  accommodate these kinds of actions, we consider each action sequence $\mathcal{X}$ to have a decoupled global component and a local component as described below. 

\noindent \textbf{Global component} ($\mathcal{X}_g$): This contains global trajectory sequence of the root node associated with per-timestep kinematic pose tree on a per-person basis. Instead of maintaining independent per-person global trajectories, we model relative displacements of individuals with respect to the first person  (see `Global component' in Fig.~\ref{fig:architecture}). Let the first person's root node global trajectory be denoted as $G^{(1)} = \left[ g_1 , g_2,\ldots \right]$ where  $g_i \in \mathbb{R}^{3}$. Let the relative displacement sequence for the j-th person's root node ($1 < j \leqslant P$) be $D^{(j)} = \left[ d_1 , d_2,\ldots \right]$ where  $d_i \in \mathbb{R}^{3}$. Thus, the global trajectory for $j$-th person's root node is given by $G^{(j)} = G^{(1)} + D^{(j)}$. Note that $G^{(1)}$ and $D^{(j)}, 1 < j \leqslant P$ together comprise the global component $\mathcal{X}_g$.

\noindent \textbf{Local component} ($\mathcal{X}_l$): This comprises of per-person kinematic pose tree sequences whose per-timestep kinematic tree root nodes are all considered to be at the global origin (see Fig.~\ref{fig:architecture}). Representing the pose tree with local 3-D position based joint representations seems a straightforward choice. However, in practice, this generates unnatural-looking motion arising from non-constrained bone length and joint movement beyond normal articulation range. To overcome these issues, we use a fixed reference pose and forward kinematics to model the pose tree's relative joint displacements as a rotation matrix. Using the procedure of Zhou et.al.~\cite{rot6d} (purple shaded box in Fig.~\ref{fig:architecture}), we adopt a continuous 6-D representation for the rotation matrix. This choice enables us to bypass restrictive post-processing required to maintain orthogonality of the usual $3 \times 3$ joint rotation matrix. Consequently, we model the local pose sequences comprising the action as $\mathcal{X}_l = \{[X_l^{(1)}, X_l^{(2)},\ldots X_l^{(p)}]_t\}$ where $[X_l^{(i)}]_t \in \mathbb{R}^{J \times \mathbf{6}}$, i.e. 6-D rotation representation of $J$ joints, $1 \leqslant t \leqslant T$. For sequences with a single person, the reference sequence is duplicated $P$ times for consistent processing.

\section{Our approach (MUGL)}
\label{sec:architecture}

MUGL comprises of three encoder modules to represent the local component of action sequence, the corresponding global component and the sequence length. Corresponding to the these modules, three decoder counterpart modules exist (see Fig.~\ref{fig:architecture}). We first provide a brief overview of our latent generative framework. We subsequently describe the previously mentioned modules (Sec.~\ref{sec:vaeencoder}, \ref{sec:vaedecoder}) and the associated latent representation (Sec.~\ref{sec:latentspace}).  

\subsection{Conditional Gaussian Mixture VAE}
\label{sec:cgmvae}

In a traditional Variational Auto Encoder (VAE)\cite{VAE} generative model, we maximize the so-called variational lower bound (ELBO) i.e. $\mathcal{L}(\theta,\phi;\mathcal{X}) = \mathbb{E}_{q_\phi(z|\mathcal{X})}[\log p_\theta(\mathcal{X}|z)] - D_{KL}(q_\phi(z|\mathcal{X})||p(z))$, where $q_\phi(z|\mathcal{X})$ encodes  variational approximation of the latent surrogate $z$'s distribution conditioned on input data $\mathcal{X}$ while $p_\theta(\mathcal{X}|z)$ approximates the latent-conditioned data likelihood. $D_{KL}$ stands for KL-divergence. A popular choice for latent prior distribution $p(z)$ is the unimodal standard Gaussian, i.e. $p(z) = \mathcal{N}(0,I)$ because it enables a closed-form solution for computing KL-divergence. However, the unimodality can become a capacity bottleneck especially when data is expected to contain a large number of clusters due to attribute-based (e.g. category, action dynamics) similarity among samples. Therefore, we adopt an extension of VAE known as Gaussian Mixture VAE (GMVAE)~\cite{GMVAE} wherein the latent prior is a Mixture of Gaussian. Furthermore, we use a conditioning strategy to incorporate additional action attribute information as part of the inference and generation process. We dub the resulting model C-GMVAE. In this framework, the encoder models $q_\phi(y,z|\mathcal{X},\mathbf{a})$ while the decoder models $p_\theta(\mathcal{X}|y,z,\mathbf{a})$ where $\mathbf{a}$ represents conditioning information (action class $c$ and viewpoint $v$). $y$ denotes the weight distribution over Gaussian components of the mixture. The ELBO in this case can be written as:
\begin{equation} 
\begin{array}{l}
    \mathcal{L}(\theta,\phi,\mathbf{a};\mathcal{X}) = \mathbb{E}_{q_\phi(y,z|\mathcal{X},\mathbf{a})}[\log p_\theta(\mathcal{X}|y,z,\mathbf{a})] \\ 
    - D_{KL}(q_\phi(y,z|\mathcal{X},\mathbf{a})||p(y,z|\mathbf{a}))
\end{array}
\label{eqn:cgmvae-elbo}
\end{equation}

\subsection{Encoder Modules}
\label{sec:vaeencoder}

Recall that we represent action sequence $\mathcal{X}$ in terms of  global component $\mathcal{X}_g$ and local component $\mathcal{X}_l$ (Sec.~\ref{sec:skelerotationrep}).

\noindent \textbf{Local Pose Encoder:} The local component $\mathcal{X}_l \in \mathbb{R}^{T \times 6 \times (J \times P)}$ is processed in two phases. The first phase (Spatial Pose Encoder) contains a single 2D residual CNN block and learns features for each timestep in $\mathcal{X}_l$. The second phase (Temporal Encoder) applies multiple 2D residual CNN convolutions to downsample the timesteps $T$. Finally, the output from the Temporal Encoder is flattened to produce the spatio-temporal local features $f_l$ (see Fig.~\ref{fig:architecture}). 

\noindent \textbf{Global Trajectory Encoder:} The global component $\mathcal{X}_g \in \mathbb{R}^{T \times 3 \times P}$ is downsampled via a series of 1D convolutions across the timestep dimension. The result is transformed by a linear layer to obtain a flattened feature representation $f_g$. Note that the dimension $P$ in $\mathcal{X}_g$ arises from global trajectory for first person and $(P-1)$ relative displacement sequences for the other $(P-1)$ people involved in the action. 

\noindent \textbf{Sequence Length Encoder:} Let $t_{s}$ denote the length of a particular training sequence. To encode the sequence length, we first define an indexed sequence $t[n]$ of length $T$ as follows:

\begin{math}
  t[n] =\left\{
    \begin{array}{ll}
      \frac{n}{t_{s}-1} & \mbox{if $0 \leqslant n < t_{s}$}.\\
      1 & \mbox{if $t_s \leqslant n < T$}.
    \end{array}
  \right.
  \label{eqn:index}
\end{math}

where $T$ is the maximum possible sequence length. Note that $t[n]$ is a normalized non-decreasing sequence of length $T$ whose values lie in $[0,1]$ (see Fig.~\ref{fig:architecture}). The Sequence Length Encoder transforms $t[n]$ to $f_s$ via 1-D convolutions.

\noindent \textbf{Encoder Representation:} The local pose sequence representation $f_l$, global trajectory sequence representation $f_g$ and sequence length representation $f_s$ are then concatenated. During training, this concatenated representation is conditionally modulated with action class-label $c$ (See Fig~\ref{fig:architecture}). In addition, we also condition on a transformed version of viewpoint $v$ to enrich the encoded representation. The result is mapped to the parametric representations of the C-GMVAE's variational approximation distribution $q_\phi(y,z|\mathcal{X},\mathbf{a}=(c,v))$ (Equation~\ref{eqn:cgmvae-elbo}). 

\subsection{Latent Representation}
\label{sec:latentspace}

We first sample a latent vector $z$ from the mixture of $K$ Gaussian components (Sec.~\ref{sec:cgmvae}). Similar to the conditioning applied on the encoder representation, the latent vector $z$ is conditioned on class label and transformed viewpoint, by concatenation (see Fig.~\ref{fig:architecture}). The resulting is transformed by a linear layer and reshaped to obtain $z_{cv}$. Note that the training sequence's viewpoint is used for conditioning only during training. During generation, the viewpoint is set to a fixed default value.

\begin{table*}[!t]
\centering
\resizebox{\textwidth}{!}
{
\centering
\begin{tabular}{c|ccccrccc}
\toprule
     Model & Class & Support for & Variable sequence & Input & \# parameters & Inference & Training \\
           & conditioning & multi-person actions & length generation &        &              & time (sec)   & time (hrs) \\
    \midrule
    VAE-LSTM\cite{vaelstm} & \xmark & \xmark & \xmark & 3-D joints & $ 1468 K$ & $ 0.03 $ & $ 45.14 $ \\
    CS-GCN\cite{CSGCN} &  \xmark & \xmark & \xmark & 3-D joints & $ 7744 K$ & $ 0.60 $ & $ 10.74 $ \\
    SA-GCN\cite{yu2020structureaware} & \cmark ($10$) & \xmark & \xmark & 2-D joints & $ 14502 K$ & $ 3.31 $ & $ 71.90 $\\
    action2motion\cite{action2motion} & \cmark ($13$) & \xmark & \xmark &  Lie Space & $503 K$ & $ 3.31 $ & $ 40.27 $\\
    \midrule
    \textbf{MUGL (ours)} & \cmark (\textbf{120}) & \textbf{\cmark} & \textbf{\cmark} & \textbf{6-D\cite{rot6d}} & $ \mathbf{922 K} $ & $ \textbf{0.02} $ & $ \textbf{18.92} $\\
    \bottomrule
\end{tabular}
 }
\captionof{table}{A comparative summary of  models used for evaluation.}
\label{tab:modelsstats}
\end{table*}

\begin{table}[!t]
\resizebox{\linewidth}{!}
{
\centering
\begin{tabular}{c|c|c|ccc|ccccc}
\toprule
     \# Classes & \# Viewpoint & Frame rate & \multicolumn{3}{c|}{Sequence length} & \multicolumn{4}{c}{Hyperparameters}\\
     & & & Max & Min & Avg & $\lambda_{rot}$  & $\lambda_{global}$ & $\lambda_{len}$ & Batch size & $\theta_{s}$ \\
    \midrule
    $120$ & $3$ & $8.25$ & $256$ & $5$ & $69.72$ & $10$ & $1$ & $2$ & $100$ & $0.97$ \\
    \bottomrule
\end{tabular}
 }
\captionof{table}{Salient dataset attributes and hyperparameter choices.}
\label{tab:datasetsstats}
\end{table}

\subsection{Decoders}
\label{sec:vaedecoder}

\noindent \textbf{Sequence Decoder:}  Complementary to encoding phase (Sec.~\ref{sec:vaeencoder}), the Local Spatial and Temporal Decoders perform spatiotemporal upsampling. Starting from the conditioned latent representation $z_{cv}$, the local pose components of generated action sequence with 6-D rotation representations for each joint are obtained. Instead of transpose convolutions in the local Spatial Decoder (mirroring their Spatial Encoder counterparts), we use a multi-layer perceptron. Empirically, we found this to generate better quality sequences. 

\noindent \textbf{Forward Kinematics:} Similar to \cite{SMPL} this module (shown as purple shaded box in Fig.~\ref{fig:architecture}) computes 3-D positions of each joint from the local representation output by Temporal Decoder. First, it converts the 6-D rotation representation into $3 \times 3$ rotation matrix corresponding to every joint, i.e. $\widetilde{\mathcal{X}_l}  =  \{[X_l^{(1)}, X_l^{(2)}, ..., X_l^{(p)}]_t\}$ where $X_l^{(i)} \in \mathbb{R}^{J \times 3 \times 3}$. The forward kinematic function then takes $\widetilde{\mathcal{X}_l}$, vertices of rest pose $\widetilde{\mathcal{R}_l}$ (see Fig.~\ref{fig:architecture}) and the kinematic tree as input. It applies transformation $F_{fkt}$ which outputs 3-D positions of joints $\widetilde{\mathcal{X}}_e = \{[X_e^{(1)}, X_e^{(2)}, \ldots X_e^{(p)}]_t\}$, where pose instance $X_e^{(i)} \in \mathbb{R}^{J \times 3}$. The transformation $F_{fkt}$ is given as:

\begin{math}
  F_{fkt}(X_l^{{(i)}^{(c)}}, \widetilde{\mathcal{R}}) =\left\{
    \begin{array}{ll}
      [0, 0, 0] & \mbox{if root joint}.\\
      X_l^{{(i)}^{(c)}} \cdot (\widetilde{\mathcal{R}}^{c} -  \widetilde{\mathcal{R}}^{p}) \\
      + F_{fkt}(X_l^{{(i)}^{(p)}}, \widetilde{\mathcal{R}}) & \mbox{otherwise}.
    \end{array}
  \right.
  \label{eqn:forward_kinematics}
\end{math}

where $X_l^{{(i)}^{(c)}}$ and $X_l^{{(i)}^{(p)}}$ indicate $3 \times 3$ rotation matrix of child and parent joints respectively in the kinematic tree of pose instance $X_l^{(i)}$. $\widetilde{\mathcal{R}}^{c}$ and $\widetilde{\mathcal{R}}^{p}$ indicate 3D joint position of child and parent joints respectively in the kinematic tree of rest pose $\widetilde{\mathcal{R}}$. Finally, the 3-D joint positions of pose instance $X_e^{(i)} \in \mathbb{R}^{J \times 3}$ is given as, $X_e^{(i)} = [f_{fkt}(X_l^{{(i)}^{(1)}}, \widetilde{\mathcal{R}}), f_{fkt}(X_l^{{(i)}^{(2)}}, \widetilde{\mathcal{R}}), ..., f_{fkt}(X_l^{{(i)}^{(J)}}, \widetilde{\mathcal{R}})]$.

This process ensures consistency in the bone lengths throughout the sequence. Note that $\widetilde{\mathcal{X}}_e$ represents local pose structure sequence, i.e. joint positions considering root joint of pose kinematic tree (torso) to be at the global origin for each time step. Also, the transformation from 6D space to $3 \times 3$ rotation matrix space is continuous and the forward kinematic function is differentiable with respect to 3-D rotation. This enables incorporation of $\widetilde{\mathcal{X}_l}$ into the overall optimization procedure (Sec.~\ref{sec:optimization}). 

Finally, note that the `Inverse Kinematics' module in the encoder portion of MUGL performs the opposite of the process described above, i.e. it maps the 3-D joint based local pose sequence to the 6-D rotation based counterpart. 

\noindent \textbf{Global Trajectory Decoder:} The conditioned latent sample $z_{cv}$ is gradually upsampled via a series of 1-D CNN layers to a $T$ timestep sequence. The resulting sequence is transformed by a linear layer to generate the global trajectory $\widetilde{\mathcal{X}}_g \in \mathbb{R}^{T \times P \times 3}$ of the reference (first) person and  relative displacements of the remaining $(P-1)$ people involved in the activity (Sec.~\ref{sec:vaeencoder}). 

The local pose sequences $\widetilde{\mathcal{X}}_e$ are appropriately offset using the information from global trajectory decoder $\widetilde{\mathcal{X}}_g$ to obtain the final generated sequence $\widetilde{\mathcal{X}}$ (see Fig.~\ref{fig:architecture}). Since our procedure uniformly generates sequences for $P$ individuals, we consider only the first person's sequence for single person action classes. 

\noindent \textbf{Sequence length Decoder:} The latent representation $z_{cv}$ is transformed via a linear layer and the resulting output is processed via 1-D convolutions to obtain a $\mathbb{R}^{T \times 1}$ sequence. This sequence is transformed by a non-negative activation (ReLU). Subsequently, the cumulative sum sequence of the activation transformed sequence is obtained. This ensures that the resulting sequence is non-decreasing. The sequence elements are then transformed via a sigmoid activation to obtain $\widehat{t[n]}$, the decoder analogue of the indexed sequence (Eqn.~\ref{eqn:index}). The location of first element in $\widehat{t[n]}$ that is greater than or equal to a fixed threshold $\theta_{s}$ is considered the length of the sequence $\widehat{t_s}$. More precisely, $\widehat{t_s} = 1+ \underset{j}{\mathrm{argmin}} \; [t[j] \geqslant \theta_{s} ], 0 \leqslant j < T$. 

\subsection{Optimization}
\label{sec:optimization}

Optimizing the overall VAE-based framework requires a tradeoff between reconstruction loss for training data and distribution approximation loss for latent space (Equation~\ref{eqn:cgmvae-elbo}). The Sequence Decoder generates joint rotation representation in 6-D space which is integrated with Forward Kinematics module to obtain 3-D joint coordinates. During optimization, the (MSE) loss for 6-D rotation representation space $\mathcal{L}_{\text{6D}}$ 
and 3-D space $L_{\text{3D}}$ together comprise the local motion reconstruction loss $\mathcal{L}^{\text{rec}}_{\text{local}} = \lambda_{rot} \mathcal{L}_{\text{6D}} + \mathcal{L}_{\text{3D}}$ where $\lambda_{rot}$ is a tradeoff hyperparameter. We combine $\mathcal{L}^{\text{rec}}_{\text{local}}$ with global trajectory (MSE) loss $\mathcal{L}^{\text{rec}}_{\text{global}}$, sequence length (MSE) loss $\mathcal{L}^{\text{rec}}_{\text{len}}$ and the KL-divergence loss of GMVAE, $\mathcal{L}_{\text{KL}}$. The effective final loss function is defined as:

\begin{equation} 
    \mathcal{L} = (\mathcal{L}^{\text{rec}}_{\text{local}} + \lambda_{global}\mathcal{L}^{\text{rec}}_{\text{global}}) + \lambda_{len}\mathcal{L}^{\text{rec}}_{\text{len}} + \lambda_{KL} \mathcal{L}_{\text{KL}}
    \label{eqn:finalloss}
\end{equation}

\noindent where $\lambda_{rot}$, $\lambda_{global}$, $\lambda_{len}$ and $\lambda_{KL}$ are hyperparameters. Note that the local and global losses are computed only for the original, non-padded extent of the training sequences.

\subsection{Implementation}

We train MUGL architecture using Adam optimizer with initial learning rate $0.015$, integrated with a learning scheduler which decays the learning by $0.5$ with step size $10$ and train the model for $200$ epochs. The value of hyper-parameters in the loss function (Equation~\ref{eqn:finalloss}) are empirically set  and the hyper-parameter for KL-divergence $\lambda_{KL}$ is determined by a cyclic annealing schedule~\cite{cyclic_annealing}.

Although the original frame rate of NTU-RGBD is 33.33 fps, such high resolution is not required for training. Therefore, we subsample the sequences by a factor of $4$ (See Table~\ref{tab:datasetsstats}).

During inference, we sample from the mixture of $K=|\mathcal{C}|$ Gaussian components (i.e. $K$ is same as number of action classes). The latent vector is conditioned on one-hot representations of desired action class. The resulting vector is fed in parallel to local and global components of the Sequence Decoder. The local component generates an action sequence in the form of 6D rotation representation. The representation is transformed to its $3 \times 3$ counterpart and together with the reference rest pose, forward kinematics is applied to obtain the local pose. This is combined with the output from Global Trajectory Decoder module to obtain the initial action sequence. The effective sequence length obtained from Sequence Length Decoder (Sec.~\ref{sec:vaedecoder}) is used to trim the initial action sequence to the final result. We perform class distribution based oversampling for minibatches to compensate for the relatively small number of action classes involving leg movement in our dataset. To match the ground truth data's frame rate we increase MUGL's output sequence resolution in the temporal dimension from $64$ timesteps to $256$ using bicubic interpolation.

We conduct all experiments on a machine with an Intel Xeon E5-2640 v4 and Nvidia GeForce GTX 1080 Ti 11GB GPUs with Ubuntu 16.04 OS. Our code uses Python-3.7 and PyTorch-0.4 library. See Table ~\ref{tab:datasetsstats} for hyperparameter settings.

\begin{figure*}[!t]
    \includegraphics[width=\textwidth]{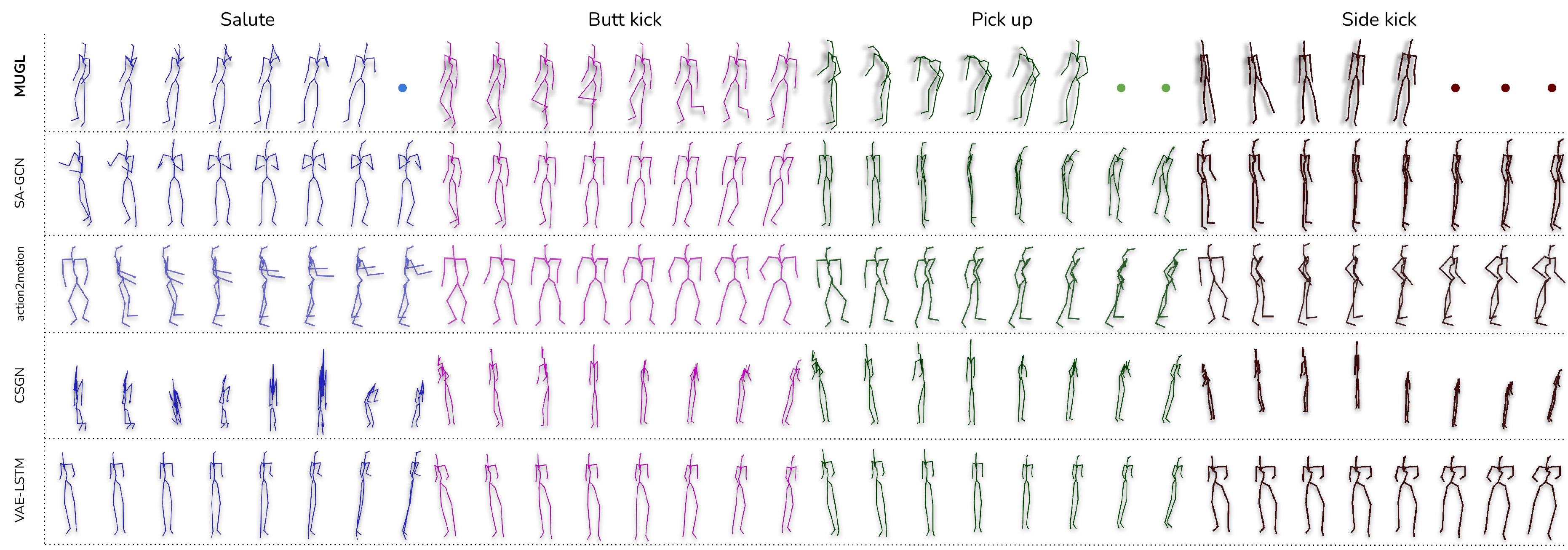}
    \caption{Comparing conditionally generated single-person action sequences across models. Also, note the variable sequence length of MUGL's examples. (ref. Sec.~\ref{sec:results}).}
    \label{fig:generations}
\end{figure*}

\section{Experiments}

\subsection{Hybrid Pose Sequence Representation}
\label{sec:ntuvibe120}

The original NTU dataset~\cite{ntu-120} contains Kinect-based 3D skeleton data with temporally incoherent, noisy and uncurated joint data. Na\"ively using this data for training causes models to generate poor quality action sequences. Therefore, we use VIBE~\cite{vibe}, a state of the art RGB-based 3D pose estimator to obtain skeleton sequences from video sequences of NTU-RGBD-120 dataset~\cite{ntu-120}. Unlike Kinect, VIBE utilizes the complete video to estimate 3D pose with minimal variance in bone lengths across the sequence. The obtained pose estimates are much more reliable due to additional context-based filtering. Since VIBE only provides local pose (See.\ref{sec:skelerotationrep}), we use corresponding samples from the original NTU-120 Kinect sequence to obtain the global trajectory of each subject. For multi-person activities, associating the per-person global trajectories to the VIBE-based pose counterparts is done based on similarity in each subject's orientation.

We ultimately produce hybrid pose sequences spanning the $120$ action classes and $114{,}480$ samples. We preprocess sequences (padding, cropping) to ensure a uniform length of $256$ timesteps. For training and evaluation, we follow the cross-setup protocol defined for NTU-RGBD-120~\cite{ntu-120}. In this protocol, action sequences from half of the total camera setups are used for training and the sequences captured from the other half are used for evaluation.

\begin{figure}[!t]
\centering
    \includegraphics[width=0.5\textwidth]{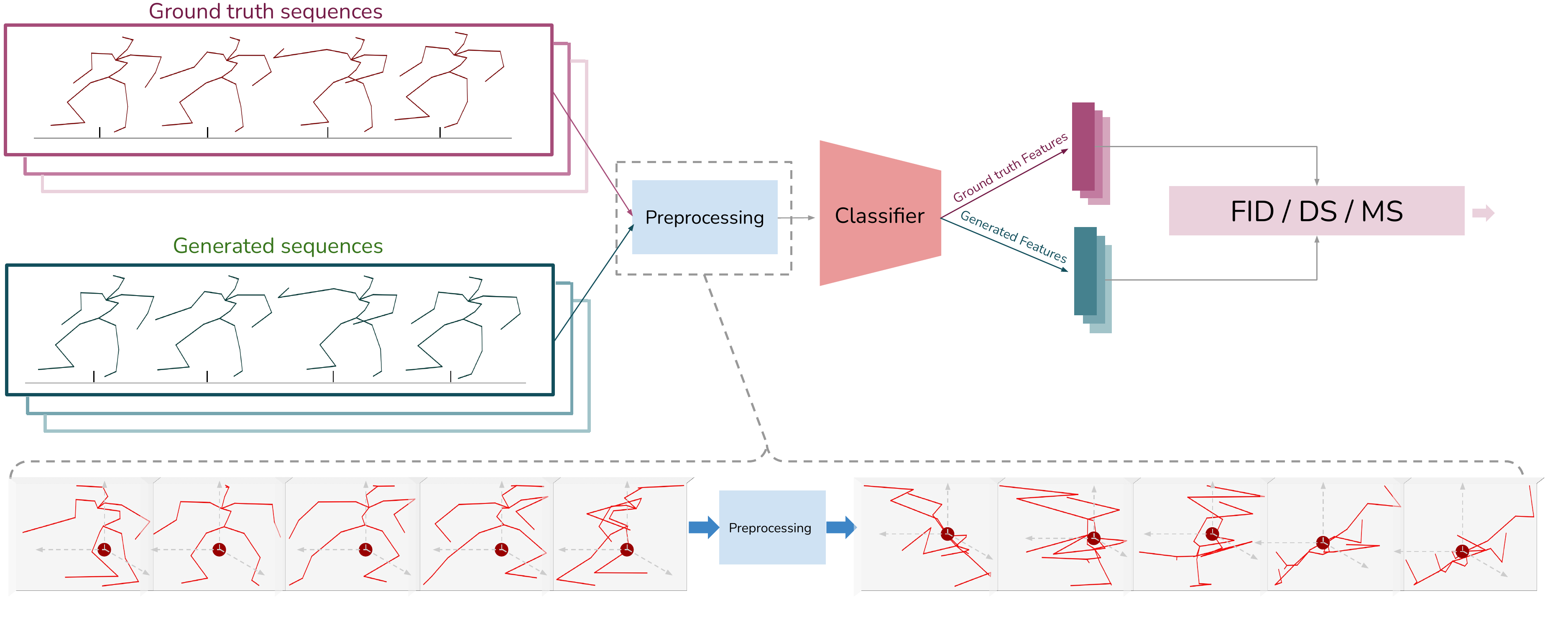}
    \caption{The typical pipeline for computing feature representation based generative quality scores is shown. The sequence at bottom right shows the effect of preprocessing. For e.g., even though rotation (about vertical axis) might be a signature of original action,  we see that preprocessing can distort and eliminate such signature components. Quality score (e.g. FID) based on feature representations of such sequences fail to capture the key action dynamics. We empirically observed that these scores correlate poorly with visual quality of category-conditioned action generations.}
    \label{fig:preprocessing}
\end{figure}

\subsection{Baselines}

We compare MUGL with four baseline generative models, namely SA-GCN~\cite{yu2020structureaware}, action2motion~\cite{action2motion}, CS-GCN~\cite{CSGCN} and VAE-LSTM~\cite{vaelstm} (see Table~\ref{tab:modelsstats}). We use available implementations for the baselines except for CS-GCN~\cite{CSGCN}, which we implement from scratch and extend by adding action class conditioning. We perform a similar implementation-based extension for VAE-LSTM~\cite{vaelstm}. 

\subsection{Generation Quality Measures}
\label{sec:metrics}

To measure the quality of generated sequences, we use two variants of Maximum mean discrepancy (MMD)~\cite{MMD}. 

\noindent \textbf{MMD-A:} Maximum Mean Discrepancy captures similarity between generated and test set (ground-truth) sample distributions~\cite{MMD,yu2020structureaware,Battan_2021_WACV}. For our setting (MMD-A), the base similarity is measured on a per-timestep basis for sequence pairs $g,e$ sampled from generated set $G$ and test set $E$. Let $g_t \in \mathbb{R}^{J \times 3}$ and $e_t \in \mathbb{R}^{J \times 3}$ represent the $t$-th timestep poses of the sampled pair and having same action class. The base similarity (MMD-A) is computed as  $\mathcal{K}(g_t,g_t) + \mathcal{K}(e_t,e_t) - 2 \mathcal{K}(g_t,e_t)$ where $\mathcal{K}$ is a similarity kernel. In particular, we employ the RBF kernel~\cite{JMLR:v11:chang10a}. 

\noindent \textbf{MMD-S:} Unlike MMD-A, MMD-S is computed on the whole sequence. Let, $g,e$ be sequences chosen from generated set $G$ and test set $E$, where $g,e \in \mathbb{R}^{T \times J \times 3}$. We flatten $g,e$ to get a vector representation of the whole sequence. MMD-S is computed as $\mathcal{K}(g,g) + \mathcal{K}(e,e) - 2 \mathcal{K}(g,e)$. 

\begin{table}[!t]
\centering
\resizebox{\linewidth}{!}
{
\centering
\begin{tabular}{r|cc}
      & \multicolumn{2}{c}{NTU-VIBE-94 (Single person classes only) }\\
\toprule
Model & MMD-A $\; \downarrow$  & MMD-S $\; \downarrow$ \\
\midrule
SA-GCN\cite{yu2020structureaware} 
 & $0.68^{\pm 0.12}$ & $0.43^{\pm 0.02}$ \\
action2motion\cite{action2motion} & $0.57^{\pm 0.11}$ & $0.52^{\pm 0.03}$ \\
CS-GCN\cite{CSGCN} & $1.09^{\pm 0.17}$ & $0.56^{\pm 0.01}$ \\
VAE-LSTM\cite{vaelstm} & $1.11^{\pm 0.17}$ & $0.54^{\pm 0.01}$ \\
\midrule
\textbf{MUGL} & $\mathbf{0.34^{\pm 0.12}}$ & $\mathbf{0.17^{\pm 0.01}}$ \\

\midrule
      & \multicolumn{2}{c}{NTU-VIBE-120}\\
\midrule

 \textbf{MUGL (Multi person)} & $\mathbf{0.45^{\pm 0.15}}$ & $\mathbf{0.36^{\pm 0.03}}$ \\

\bottomrule
\end{tabular}
 }
\captionof{table}{Model comparison in terms of generative quality scores on NTU-VIBE dataset. $\downarrow$ indicates that the motions are better when metric is lower. For fair comparison, all the baselines are trained on 94 single-person classes. Single person and multi-person variant of MUGL are not directly comparable but exhibit similar trends.}
\label{tab:comparision}
\end{table}

\noindent \textbf{Issues with feature-based generative quality measures:} In existing works, measures such as Fr\'{e}chet Inception Distance (FID)~\cite{FID}, Diversity score (DS)~\cite{action2motion} and Multimodality score (MS)~\cite{action2motion} are commonly reported. These measures rely on a pre-trained classifier's feature representation of the input (real or generated). Most skeleton action classifiers employ a preprocessing step in which root joints of pose sequence are translated to origin and the sequence is transformed so that the person shoulder faces x-axis. Doing so eliminates variability arising from camera viewpoint and enables good classification performance. However, as Fig.~\ref{fig:preprocessing} shows, the preprocessing severely distorts the original action's dynamics. Removing preprocessing would seem a possible solution. However, doing so causes a significant drop in classifier performance (to less than $20\%$), making the resulting feature representations unreliable. 

Therefore, generative quality scores (FID, MS, DS) obtained from feature representations of such preprocessed action sequences cannot be considered reliable for evaluating action sequences. 

\subsection{Results}
\label{sec:results}

To compute performance scores for a given model, we uniformly generate $300$ samples per action class. Since baseline models are confined to single-person generation and have no provision for multi-person setting, we use a single-person variant of the dataset with $94$ action classes for training and evaluation of all the models. The quantitative results can be viewed in Table~\ref{tab:comparision}. MUGL's scores are visibly better than the baselines. Empirically, MMD scores (Table.~\ref{tab:comparision}) correlate well with observed visual quality. The advantage of our approach can also be appreciated from a comparison of compute attributes for various models given in Table~\ref{tab:modelsstats}. We compute the quality measures for two-person action generation as well (see bottom row of Table~\ref{tab:comparision}). Since the model configurations (one-person and two-person) are different, the scores are not directly comparable. However, it can be seen that two-person generation scores are similar in range to the one-person variant. 

Fig.~\ref{fig:generations} provides a qualitative illustration with generated samples from some of the action categories. The results, including Fig.~\ref{fig:introfig}, demonstrate the efficacy of our approach in generating diverse action-conditioned variable-length action sequences. This is especially apparent for actions involving significant limb movement.

To explore the variability of performance due to design choices (action representation, architectural components and optimization), we computed scores for ablative variants of MUGL trained on all $120$ NTU-120 classes. From the results in Table~\ref{tab:ablations}, we observe the following: (i) performance is lower with vanilla VAE at sequence level (MMD-S) (ii) MLP in Spatial Decoder noticeably improves performance compared to the usual 2D Transpose Convolution (iii) class and viewpoint conditioning matters (iv) performance is lower with raw 3D joint representation (v) generating fixed length sequences (by removing variable sequence length encoder-decoder)  negatively affects performance. 

Overall, the results demonstrate the suitability of our design choices. The benefit of our decoupled local and global sequence modelling, the choice of CNNs for computational efficiency, importance of variable sequence estimation, the increased latent distribution capacity due to Gaussian Mixture model are all evident.

\begin{table*}[!t]
\centering
\resizebox{\linewidth}{!}
{
 \begin{tabular}{ c|c|l|c|c|c}
 \toprule
 Ablation Type & Pipeline Component & Ablation Details  & MMD-A $\;\downarrow$   & MMD-S $\;\downarrow$ \\
 \midrule
\multirow[c]{5}{*}[-2.5em]{\rule{0pt}{2ex}Architectural} & \multirow[c]{4}{*}[-0.5em]{\rule{0pt}{2ex} VAE (Sec.~\ref{sec:cgmvae})}
& Reduce VAE latent dimensions (0.5x) & $0.47$ & $0.47$ \\
& & Increase VAE latent dimensions (2x) & $0.48$ & $0.55$\\
& & Reduce GMM components ($K=60$) & $\mathbf{0.41}$ & $0.39$\\
& & Increase GMM components ($K=240$) & $0.47$ & $0.39$\\
& & Vanilla VAE: Unimodal Gaussian ($K=1$) & $0.45 $ & $0.41$\\
\cline{2-5} 
& \multirow[c]{3}{*}[-0.5em]{\rule{0pt}{2ex} Encoder \& Decoder (Sec.~\ref{sec:vaeencoder},\ref{sec:vaedecoder}) } 
& MLP in Spatial Encoder & $0.46$ & $0.39$\\
& & Transpose 2D conv in Spatial Decoder  & $0.65$ & $0.64$\\
& & No class, viewpoint conditioning in Sequence Encoder & $0.48$ & $0.41$\\
& & 3-D joint representation as input, output & $0.46$ & $0.45$\\
\cline{2-5}
& \multirow[c]{3}{*}[-0.15em]{\rule{0pt}{2ex} Viewpoint conditioning
(Sec.~\ref{sec:latentspace})} 
& Using one-hot vector/remove viewpoint transformer & $0.49$ & $0.41$\\
& & No viewpoint conditioning & $0.45$ & $0.39$\\
& & No viewpoint conditioning in latent & $0.79$ & $0.75$\\
\hline
& \multirow[c]{1}{*}[-0.1em]{\rule{0pt}{2ex} Variable sequence length} 
& Remove variable sequence length encoder-decoder & $1.77$ & $0.79$\\
\hline
\multirow{2}{*}[-0.25em]{\rule{0pt}{2ex} Optimization } & \multirow[c]{3}{*}[0.5em]{\rule{0pt}{2ex} Sequence Decoder (Sec.~\ref{sec:vaedecoder})  } 
& No 3D Loss & $0.74$ & $0.37$ \\
& & No Rotation Loss & $0.56$ & $0.39$ \\
\hline
\multicolumn{3}{c|}{\textbf{MUGL (multi-person)}} & $0.45$ & $\mathbf{0.36}$\\
\bottomrule
\end{tabular}
}
\captionof{table}{Performance scores for MUGL ablative variants.} 
\label{tab:ablations}
\end{table*}

\section{Conclusion}

We have introduced MUGL, a novel deep neural model for variable-length pose-based action generation with locomotion. Our controllable method enables diverse generation of single and multi-person actions at scale.  MUGL outperforms strong baselines  qualitatively and quantitatively in terms of generated sequences. Our generated sequences can be used for rendering, e.g. using skeleton pose as $\alpha$ and a desired shape profile as $\beta$ parameter configurations in SMPL system~\cite{SMPL}. Going forward, we intend to explore MUGL's performance on other human pose action datasets. 

\section*{Acknowledgements}

We acknowledge Qualcomm Innovation Fellowship's support to Debtanu Gupta and Google Cloud's GPU credits for academic research.

\bibliographystyle{ACM-Reference-Format}
\balance
\bibliography{main} 


\begin{thebibliography}{40}


\ifx \showCODEN    \undefined \def \showCODEN     #1{\unskip}     \fi
\ifx \showDOI      \undefined \def \showDOI       #1{#1}\fi
\ifx \showISBNx    \undefined \def \showISBNx     #1{\unskip}     \fi
\ifx \showISBNxiii \undefined \def \showISBNxiii  #1{\unskip}     \fi
\ifx \showISSN     \undefined \def \showISSN      #1{\unskip}     \fi
\ifx \showLCCN     \undefined \def \showLCCN      #1{\unskip}     \fi
\ifx \shownote     \undefined \def \shownote      #1{#1}          \fi
\ifx \showarticletitle \undefined \def \showarticletitle #1{#1}   \fi
\ifx \showURL      \undefined \def \showURL       {\relax}        \fi
\providecommand\bibfield[2]{#2}
\providecommand\bibinfo[2]{#2}
\providecommand\natexlab[1]{#1}
\providecommand\showeprint[2][]{arXiv:#2}

\bibitem[\protect\citeauthoryear{Ahn, Ha, et~al\mbox{.}}{Ahn
  et~al\mbox{.}}{2018}]%
        {text2action}
\bibfield{author}{\bibinfo{person}{Hyemin Ahn}, \bibinfo{person}{Timothy Ha},
  {et~al\mbox{.}}} \bibinfo{year}{2018}\natexlab{}.
\newblock \showarticletitle{Text2Action: Generative Adversarial Synthesis from
  Language to Action}. In \bibinfo{booktitle}{\emph{ICRA}}.
  \bibinfo{pages}{5915--5920}.
\newblock


\bibitem[\protect\citeauthoryear{Ahuja and Morency}{Ahuja and Morency}{2019}]%
        {language2pose}
\bibfield{author}{\bibinfo{person}{C. Ahuja} {and} \bibinfo{person}{L.
  Morency}.} \bibinfo{year}{2019}\natexlab{}.
\newblock \showarticletitle{Language2Pose: Natural Language Grounded Pose
  Forecasting}. In \bibinfo{booktitle}{\emph{3DV}}.
\newblock


\bibitem[\protect\citeauthoryear{Arikan}{Arikan}{2003}]%
        {motion_annotations}
\bibfield{author}{\bibinfo{person}{Okan Arikan}.}
  \bibinfo{year}{2003}\natexlab{}.
\newblock \showarticletitle{Motion Synthesis from Annotations}.
\newblock \bibinfo{journal}{\emph{ACM SIGGRAPH}} \bibinfo{volume}{22},
  \bibinfo{number}{3} (\bibinfo{year}{2003}), \bibinfo{pages}{402–408}.
\newblock


\bibitem[\protect\citeauthoryear{Arikan and Forsyth}{Arikan and
  Forsyth}{2002}]%
        {motion_examples}
\bibfield{author}{\bibinfo{person}{Okan Arikan} {and} \bibinfo{person}{D.~A.
  Forsyth}.} \bibinfo{year}{2002}\natexlab{}.
\newblock \showarticletitle{Interactive Motion Generation from Examples}. In
  \bibinfo{booktitle}{\emph{CGIT}}. \bibinfo{pages}{483–490}.
\newblock


\bibitem[\protect\citeauthoryear{Barsoum, Kender, et~al\mbox{.}}{Barsoum
  et~al\mbox{.}}{2017}]%
        {hpgan}
\bibfield{author}{\bibinfo{person}{Emad Barsoum}, \bibinfo{person}{John
  Kender}, {et~al\mbox{.}}} \bibinfo{year}{2017}\natexlab{}.
\newblock \showarticletitle{{HP-GAN:} Probabilistic 3D human motion prediction
  via {GAN}}.
\newblock \bibinfo{journal}{\emph{ArXiv}} (\bibinfo{year}{2017}).
\newblock


\bibitem[\protect\citeauthoryear{Battan, Agrawal, Rao, Goel, and Sharma}{Battan
  et~al\mbox{.}}{2021}]%
        {Battan_2021_WACV}
\bibfield{author}{\bibinfo{person}{Neeraj Battan}, \bibinfo{person}{Yudhik
  Agrawal}, \bibinfo{person}{Sai~Soorya Rao}, \bibinfo{person}{Aman Goel},
  {and} \bibinfo{person}{Avinash Sharma}.} \bibinfo{year}{2021}\natexlab{}.
\newblock \showarticletitle{GlocalNet: Class-Aware Long-Term Human Motion
  Synthesis}. In \bibinfo{booktitle}{\emph{WACV}}. \bibinfo{pages}{879--888}.
\newblock


\bibitem[\protect\citeauthoryear{Cao, Gao, Mangalam, Cai, Vo, and Malik}{Cao
  et~al\mbox{.}}{2020}]%
        {caoHMP2020}
\bibfield{author}{\bibinfo{person}{Zhe Cao}, \bibinfo{person}{Hang Gao},
  \bibinfo{person}{Karttikeya Mangalam}, \bibinfo{person}{Qizhi Cai},
  \bibinfo{person}{Minh Vo}, {and} \bibinfo{person}{Jitendra Malik}.}
  \bibinfo{year}{2020}\natexlab{}.
\newblock \showarticletitle{Long-term human motion prediction with scene
  context}.
\newblock


\bibitem[\protect\citeauthoryear{Chang, Hsieh, Chang, Ringgaard, and Lin}{Chang
  et~al\mbox{.}}{2010}]%
        {JMLR:v11:chang10a}
\bibfield{author}{\bibinfo{person}{Yin-Wen Chang}, \bibinfo{person}{Cho-Jui
  Hsieh}, \bibinfo{person}{Kai-Wei Chang}, \bibinfo{person}{Michael Ringgaard},
  {and} \bibinfo{person}{Chih-Jen Lin}.} \bibinfo{year}{2010}\natexlab{}.
\newblock \showarticletitle{Training and Testing Low-degree Polynomial Data
  Mappings via Linear SVM}.
\newblock \bibinfo{journal}{\emph{Journal of Machine Learning Research}}
  \bibinfo{volume}{11}, \bibinfo{number}{48} (\bibinfo{year}{2010}),
  \bibinfo{pages}{1471--1490}.
\newblock
\urldef\tempurl%
\url{http://jmlr.org/papers/v11/chang10a.html}
\showURL{%
\tempurl}


\bibitem[\protect\citeauthoryear{CMUMocap}{CMUMocap}{2003}]%
        {cmu_mocap}
\bibfield{author}{\bibinfo{person}{CMUMocap}.} \bibinfo{year}{2003}\natexlab{}.
\newblock \showarticletitle{CMU Graphics Lab Motion Capture Database Converted
  to FBX}.
\newblock  (\bibinfo{year}{2003}).
\newblock


\bibitem[\protect\citeauthoryear{Dilokthanakul, Mediano,
  et~al\mbox{.}}{Dilokthanakul et~al\mbox{.}}{2016}]%
        {GMVAE}
\bibfield{author}{\bibinfo{person}{Nat Dilokthanakul},
  \bibinfo{person}{Pedro~AM Mediano}, {et~al\mbox{.}}}
  \bibinfo{year}{2016}\natexlab{}.
\newblock \showarticletitle{Deep Unsupervised Clustering with Gaussian Mixture
  Variational Autoencoders}.
\newblock \bibinfo{journal}{\emph{arXiv}} (\bibinfo{year}{2016}).
\newblock


\bibitem[\protect\citeauthoryear{Fragkiadaki, Levine,
  et~al\mbox{.}}{Fragkiadaki et~al\mbox{.}}{2015}]%
        {Fragkiadaki2015RecurrentNM}
\bibfield{author}{\bibinfo{person}{Katerina Fragkiadaki},
  \bibinfo{person}{Sergey Levine}, {et~al\mbox{.}}}
  \bibinfo{year}{2015}\natexlab{}.
\newblock \showarticletitle{Recurrent network models for human dynamics}. In
  \bibinfo{booktitle}{\emph{ICCV}}. \bibinfo{pages}{4346--4354}.
\newblock


\bibitem[\protect\citeauthoryear{Fu, Li, et~al\mbox{.}}{Fu
  et~al\mbox{.}}{2019}]%
        {cyclic_annealing}
\bibfield{author}{\bibinfo{person}{Hao Fu}, \bibinfo{person}{Chunyuan Li},
  {et~al\mbox{.}}} \bibinfo{year}{2019}\natexlab{}.
\newblock \showarticletitle{Cyclical Annealing Schedule: {A} Simple Approach to
  Mitigating {KL} Vanishing}.
\newblock \bibinfo{journal}{\emph{arXiv}} (\bibinfo{year}{2019}).
\newblock


\bibitem[\protect\citeauthoryear{Guo, Zuo, et~al\mbox{.}}{Guo
  et~al\mbox{.}}{2020}]%
        {action2motion}
\bibfield{author}{\bibinfo{person}{Chuan Guo}, \bibinfo{person}{Xinxin Zuo},
  {et~al\mbox{.}}} \bibinfo{year}{2020}\natexlab{}.
\newblock \showarticletitle{Action2Motion: Conditioned Generation of 3D Human
  Motions}.
\newblock \bibinfo{journal}{\emph{ACMMM}} (\bibinfo{year}{2020}).
\newblock


\bibitem[\protect\citeauthoryear{Guo and Choi}{Guo and Choi}{2019}]%
        {skelnet}
\bibfield{author}{\bibinfo{person}{Xiao Guo} {and} \bibinfo{person}{Jongmoo
  Choi}.} \bibinfo{year}{2019}\natexlab{}.
\newblock \showarticletitle{Human motion prediction via learning local
  structure representations and temporal dependencies}. In
  \bibinfo{booktitle}{\emph{AAAI}}, Vol.~\bibinfo{volume}{33}.
  \bibinfo{pages}{2580--2587}.
\newblock


\bibitem[\protect\citeauthoryear{Habibie, Holden, et~al\mbox{.}}{Habibie
  et~al\mbox{.}}{2017}]%
        {vaelstm}
\bibfield{author}{\bibinfo{person}{Ikhsanul Habibie}, \bibinfo{person}{Daniel
  Holden}, {et~al\mbox{.}}} \bibinfo{year}{2017}\natexlab{}.
\newblock \showarticletitle{A Recurrent Variational Autoencoder for Human
  Motion Synthesis}. In \bibinfo{booktitle}{\emph{BMVC}}.
\newblock


\bibitem[\protect\citeauthoryear{Heusel, Ramsauer, Unterthiner, Nessler, and
  Hochreiter}{Heusel et~al\mbox{.}}{2017}]%
        {FID}
\bibfield{author}{\bibinfo{person}{Martin Heusel}, \bibinfo{person}{Hubert
  Ramsauer}, \bibinfo{person}{Thomas Unterthiner}, \bibinfo{person}{Bernhard
  Nessler}, {and} \bibinfo{person}{Sepp Hochreiter}.}
  \bibinfo{year}{2017}\natexlab{}.
\newblock \showarticletitle{GANs Trained by a Two Time-Scale Update Rule
  Converge to a Nash Equilibrium}.
\newblock \bibinfo{journal}{\emph{arXiv}} (\bibinfo{year}{2017}).
\newblock


\bibitem[\protect\citeauthoryear{Hinton and Salakhutdinov}{Hinton and
  Salakhutdinov}{2006}]%
        {rbm}
\bibfield{author}{\bibinfo{person}{G.E. Hinton} {and} \bibinfo{person}{R.R.
  Salakhutdinov}.} \bibinfo{year}{2006}\natexlab{}.
\newblock \showarticletitle{Reducing the Dimensionality of Data with Neural
  Networks}.
\newblock \bibinfo{journal}{\emph{Science}}  \bibinfo{volume}{313}
  (\bibinfo{date}{08} \bibinfo{year}{2006}), \bibinfo{pages}{504--7}.
\newblock


\bibitem[\protect\citeauthoryear{Holden et~al\mbox{.}}{Holden
  et~al\mbox{.}}{2016}]%
        {holden_2016}
\bibfield{author}{\bibinfo{person}{Daniel Holden} {et~al\mbox{.}}}
  \bibinfo{year}{2016}\natexlab{}.
\newblock \showarticletitle{A Deep Learning Framework for Character Motion
  Synthesis and Editing}.
\newblock \bibinfo{journal}{\emph{SIGGRAPH}} \bibinfo{volume}{35},
  \bibinfo{number}{4}, Article \bibinfo{articleno}{138} (\bibinfo{year}{2016}).
\newblock


\bibitem[\protect\citeauthoryear{Holden, Komura, and Saito}{Holden
  et~al\mbox{.}}{2017}]%
        {holden_phase}
\bibfield{author}{\bibinfo{person}{Daniel Holden}, \bibinfo{person}{Taku
  Komura}, {and} \bibinfo{person}{Jun Saito}.} \bibinfo{year}{2017}\natexlab{}.
\newblock \showarticletitle{Phase-functioned neural networks for character
  control}.
\newblock \bibinfo{journal}{\emph{ACM ToG}}  \bibinfo{volume}{36}
  (\bibinfo{date}{07} \bibinfo{year}{2017}), \bibinfo{pages}{1--13}.
\newblock


\bibitem[\protect\citeauthoryear{Ionescu, Papava, et~al\mbox{.}}{Ionescu
  et~al\mbox{.}}{2014}]%
        {h36m}
\bibfield{author}{\bibinfo{person}{Catalin Ionescu}, \bibinfo{person}{Dragos
  Papava}, {et~al\mbox{.}}} \bibinfo{year}{2014}\natexlab{}.
\newblock \showarticletitle{Human3.6M: Large Scale Datasets and Predictive
  Methods for 3D Human Sensing in Natural Environments}.
\newblock \bibinfo{journal}{\emph{IEEE TPAMI}} \bibinfo{volume}{36},
  \bibinfo{number}{7} (\bibinfo{year}{2014}), \bibinfo{pages}{1325--1339}.
\newblock


\bibitem[\protect\citeauthoryear{Kingma and Welling}{Kingma and
  Welling}{2019}]%
        {VAE}
\bibfield{author}{\bibinfo{person}{Diederik~P. Kingma} {and}
  \bibinfo{person}{Max Welling}.} \bibinfo{year}{2019}\natexlab{}.
\newblock \showarticletitle{An Introduction to Variational Autoencoders}.
\newblock \bibinfo{journal}{\emph{arXiv}} (\bibinfo{year}{2019}).
\newblock


\bibitem[\protect\citeauthoryear{Kocabas, Athanasiou, et~al\mbox{.}}{Kocabas
  et~al\mbox{.}}{2020}]%
        {vibe}
\bibfield{author}{\bibinfo{person}{Muhammed Kocabas}, \bibinfo{person}{Nikos
  Athanasiou}, {et~al\mbox{.}}} \bibinfo{year}{2020}\natexlab{}.
\newblock \showarticletitle{VIBE: Video Inference for Human Body Pose and Shape
  Estimation}. In \bibinfo{booktitle}{\emph{CVPR}}.
\newblock


\bibitem[\protect\citeauthoryear{Kovar}{Kovar}{2002}]%
        {motion_graph}
\bibfield{author}{\bibinfo{person}{Lucas Kovar}.}
  \bibinfo{year}{2002}\natexlab{}.
\newblock \showarticletitle{Motion Graphs}.
\newblock \bibinfo{journal}{\emph{ACM SIGGRAPH}} \bibinfo{volume}{21},
  \bibinfo{number}{3} (\bibinfo{year}{2002}), \bibinfo{pages}{473–482}.
\newblock


\bibitem[\protect\citeauthoryear{Kundu, Gor, et~al\mbox{.}}{Kundu
  et~al\mbox{.}}{2019}]%
        {bihmpgan}
\bibfield{author}{\bibinfo{person}{Jogendra~Nath Kundu},
  \bibinfo{person}{Maharshi Gor}, {et~al\mbox{.}}}
  \bibinfo{year}{2019}\natexlab{}.
\newblock \showarticletitle{BiHMP-GAN: Bidirectional 3D Human Motion Prediction
  GAN}.
\newblock \bibinfo{journal}{\emph{AAAI}} (\bibinfo{year}{2019}).
\newblock


\bibitem[\protect\citeauthoryear{Li, Zhou, et~al\mbox{.}}{Li
  et~al\mbox{.}}{2018}]%
        {aclstm}
\bibfield{author}{\bibinfo{person}{Zimo Li}, \bibinfo{person}{Yi Zhou},
  {et~al\mbox{.}}} \bibinfo{year}{2018}\natexlab{}.
\newblock \showarticletitle{Auto-conditioned recurrent networks for extended
  complex human motion synthesis}. In \bibinfo{booktitle}{\emph{ICLR}}.
\newblock


\bibitem[\protect\citeauthoryear{Lin and Amer}{Lin and Amer}{2018}]%
        {DVGAN}
\bibfield{author}{\bibinfo{person}{Xiao Lin} {and} \bibinfo{person}{Mohamed~R.
  Amer}.} \bibinfo{year}{2018}\natexlab{}.
\newblock \showarticletitle{Human Motion Modeling using DVGANs}.
\newblock  (\bibinfo{year}{2018}).
\newblock


\bibitem[\protect\citeauthoryear{Liu, Shahroudy, et~al\mbox{.}}{Liu
  et~al\mbox{.}}{2020}]%
        {ntu-120}
\bibfield{author}{\bibinfo{person}{Jun Liu}, \bibinfo{person}{Amir Shahroudy},
  {et~al\mbox{.}}} \bibinfo{year}{2020}\natexlab{}.
\newblock \showarticletitle{NTU RGB+D 120: A Large-Scale Benchmark for 3D Human
  Activity Understanding}.
\newblock \bibinfo{journal}{\emph{TPAMI}} \bibinfo{volume}{42},
  \bibinfo{number}{10} (\bibinfo{year}{2020}), \bibinfo{pages}{2684–2701}.
\newblock


\bibitem[\protect\citeauthoryear{Loper, Mahmood, et~al\mbox{.}}{Loper
  et~al\mbox{.}}{2015}]%
        {SMPL}
\bibfield{author}{\bibinfo{person}{Matthew Loper}, \bibinfo{person}{Naureen
  Mahmood}, {et~al\mbox{.}}} \bibinfo{year}{2015}\natexlab{}.
\newblock \showarticletitle{SMPL: A Skinned Multi-Person Linear Model}.
\newblock \bibinfo{journal}{\emph{SIGGRAPH Asia}} \bibinfo{volume}{34},
  \bibinfo{number}{6} (\bibinfo{year}{2015}), \bibinfo{pages}{248:1--248:16}.
\newblock


\bibitem[\protect\citeauthoryear{Martinez, Black, et~al\mbox{.}}{Martinez
  et~al\mbox{.}}{2017}]%
        {martinez2017human}
\bibfield{author}{\bibinfo{person}{Julieta Martinez},
  \bibinfo{person}{Michael~J Black}, {et~al\mbox{.}}}
  \bibinfo{year}{2017}\natexlab{}.
\newblock \showarticletitle{On human motion prediction using recurrent neural
  networks}. In \bibinfo{booktitle}{\emph{CVPR}}. \bibinfo{pages}{2891--2900}.
\newblock


\bibitem[\protect\citeauthoryear{Pavllo, Grangier, et~al\mbox{.}}{Pavllo
  et~al\mbox{.}}{2018}]%
        {quaternet}
\bibfield{author}{\bibinfo{person}{Dario Pavllo}, \bibinfo{person}{David
  Grangier}, {et~al\mbox{.}}} \bibinfo{year}{2018}\natexlab{}.
\newblock \showarticletitle{QuaterNet: A Quaternion-based Recurrent Model for
  Human Motion}. In \bibinfo{booktitle}{\emph{BMVC}}.
\newblock


\bibitem[\protect\citeauthoryear{Peng, Abbeel, Levine, and van~de Panne}{Peng
  et~al\mbox{.}}{2018a}]%
        {deepMimic}
\bibfield{author}{\bibinfo{person}{Xue~Bin Peng}, \bibinfo{person}{Pieter
  Abbeel}, \bibinfo{person}{Sergey Levine}, {and} \bibinfo{person}{Michiel
  van~de Panne}.} \bibinfo{year}{2018}\natexlab{a}.
\newblock \showarticletitle{DeepMimic: Example-guided Deep Reinforcement
  Learning of Physics-based Character Skills}.
\newblock \bibinfo{journal}{\emph{ACM SIGGRAPH}} \bibinfo{volume}{37},
  \bibinfo{number}{4}, Article \bibinfo{articleno}{143} (\bibinfo{year}{2018}),
  \bibinfo{numpages}{143:1--143:14}~pages.
\newblock


\bibitem[\protect\citeauthoryear{Peng, Kanazawa, Malik, Abbeel, and
  Levine}{Peng et~al\mbox{.}}{2018b}]%
        {10.1145/3272127.3275014}
\bibfield{author}{\bibinfo{person}{Xue~Bin Peng}, \bibinfo{person}{Angjoo
  Kanazawa}, \bibinfo{person}{Jitendra Malik}, \bibinfo{person}{Pieter Abbeel},
  {and} \bibinfo{person}{Sergey Levine}.} \bibinfo{year}{2018}\natexlab{b}.
\newblock \showarticletitle{SFV: Reinforcement Learning of Physical Skills from
  Videos}.
\newblock \bibinfo{journal}{\emph{ACM SIGGRAPH}}, Article
  \bibinfo{articleno}{178} (\bibinfo{year}{2018}).
\newblock


\bibitem[\protect\citeauthoryear{Plappert, Mandery, and Asfour}{Plappert
  et~al\mbox{.}}{2016}]%
        {kit_motion}
\bibfield{author}{\bibinfo{person}{Matthias Plappert},
  \bibinfo{person}{Christian Mandery}, {and} \bibinfo{person}{Tamim Asfour}.}
  \bibinfo{year}{2016}\natexlab{}.
\newblock \showarticletitle{The {KIT} Motion-Language Dataset}.
\newblock \bibinfo{journal}{\emph{Big Data}} \bibinfo{volume}{4},
  \bibinfo{number}{4} (\bibinfo{year}{2016}), \bibinfo{pages}{236--252}.
\newblock


\bibitem[\protect\citeauthoryear{Sutskever and Hinton}{Sutskever and
  Hinton}{2007}]%
        {temploral_rbms}
\bibfield{author}{\bibinfo{person}{I. Sutskever} {and} \bibinfo{person}{G.
  Hinton}.} \bibinfo{year}{2007}\natexlab{}.
\newblock \showarticletitle{Learning multilevel distributed representations for
  high-dimensional sequences}. In \bibinfo{booktitle}{\emph{AISTATS}},
  Vol.~\bibinfo{volume}{2}.
\newblock


\bibitem[\protect\citeauthoryear{Sutskever, Hinton, et~al\mbox{.}}{Sutskever
  et~al\mbox{.}}{2008}]%
        {recurent_rbms}
\bibfield{author}{\bibinfo{person}{Ilya Sutskever}, \bibinfo{person}{Geoffrey~E
  Hinton}, {et~al\mbox{.}}} \bibinfo{year}{2008}\natexlab{}.
\newblock \showarticletitle{The Recurrent Temporal Restricted Boltzmann
  Machine}. In \bibinfo{booktitle}{\emph{NIPS}}. \bibinfo{pages}{1601–1608}.
\newblock


\bibitem[\protect\citeauthoryear{Tolstikhin, Sriperumbudur, and
  Sch{\"o}lkopf}{Tolstikhin et~al\mbox{.}}{2016}]%
        {MMD}
\bibfield{author}{\bibinfo{person}{Ilya~O Tolstikhin},
  \bibinfo{person}{Bharath~K Sriperumbudur}, {and} \bibinfo{person}{Bernhard
  Sch{\"o}lkopf}.} \bibinfo{year}{2016}\natexlab{}.
\newblock \showarticletitle{Minimax estimation of maximum mean discrepancy with
  radial kernels}.
\newblock \bibinfo{journal}{\emph{NIPS}} (\bibinfo{year}{2016}),
  \bibinfo{pages}{1938--1946}.
\newblock


\bibitem[\protect\citeauthoryear{Yan, Li, et~al\mbox{.}}{Yan
  et~al\mbox{.}}{2019}]%
        {CSGCN}
\bibfield{author}{\bibinfo{person}{Sijie Yan}, \bibinfo{person}{Zhizhong Li},
  {et~al\mbox{.}}} \bibinfo{year}{2019}\natexlab{}.
\newblock \showarticletitle{Convolutional Sequence Generation for
  Skeleton-Based Action Synthesis}.
\newblock  (\bibinfo{year}{2019}).
\newblock


\bibitem[\protect\citeauthoryear{Yu, Zhao, et~al\mbox{.}}{Yu
  et~al\mbox{.}}{2020}]%
        {yu2020structureaware}
\bibfield{author}{\bibinfo{person}{Ping Yu}, \bibinfo{person}{Yang Zhao},
  {et~al\mbox{.}}} \bibinfo{year}{2020}\natexlab{}.
\newblock \showarticletitle{Structure-aware human-action generation}. In
  \bibinfo{booktitle}{\emph{ECCV}}. \bibinfo{pages}{18--34}.
\newblock


\bibitem[\protect\citeauthoryear{Zhang}{Zhang}{2012}]%
        {zhang2012microsoft}
\bibfield{author}{\bibinfo{person}{Zhengyou Zhang}.}
  \bibinfo{year}{2012}\natexlab{}.
\newblock \showarticletitle{Microsoft Kinect Sensor and Its Effect}.
\newblock \bibinfo{journal}{\emph{IEEE MultiMedia}}  \bibinfo{volume}{19}
  (\bibinfo{year}{2012}), \bibinfo{pages}{4--12}.
\newblock


\bibitem[\protect\citeauthoryear{Zhou, Barnes, et~al\mbox{.}}{Zhou
  et~al\mbox{.}}{2019}]%
        {rot6d}
\bibfield{author}{\bibinfo{person}{Yi Zhou}, \bibinfo{person}{Connelly Barnes},
  {et~al\mbox{.}}} \bibinfo{year}{2019}\natexlab{}.
\newblock \showarticletitle{On the continuity of rotation representations in
  neural networks}. In \bibinfo{booktitle}{\emph{CVPR}}.
  \bibinfo{pages}{5745--5753}.
\newblock


\end{thebibliography}

\end{document}